\documentclass[fleqn,10pt]{wlscirep}
\pdfoutput=1
\usepackage[utf8]{inputenc}
\usepackage[T1]{fontenc}

\title{Gravitational Models Explain Shifts on Human Visual Attention}

\author[1,*]{Dario Zanca}
\author[2,3]{Marco Gori}
\author[2]{Stefano Melacci}
\author[1]{Alessandra Rufa}
\affil[1]{University of Siena, Department of Medicine, Surgery and Neuroscience, Siena, 53100, Italy\\ \{dario.zanca, alessandra.rufa\}@unisi.it}
\affil[2]{University of Siena, Department of Information Engineering and Mathematics, Siena, 53100, Italy\\ \{marco,mela\}@diism.unisi.it,}
\affil[3]{Université Côte d’Azur, Inria, CNRS, I3S, Maasai, Cote d’Azur, France}

\affil[*]{Corresponding author.}

\def\zanca#1{{\color{black}#1}}

\def\mela#1{{\color{black}#1}}

\def\newrev#1{{\color{black}#1}}

\begin{abstract}
Visual attention refers to the human brain's ability to select relevant sensory information for preferential processing, improving performance in visual and cognitive tasks. It proceeds in two phases. One in which visual feature maps are acquired and processed in parallel. Another where the information from these maps is merged in order to select a single location to be attended for further and more complex computations and reasoning. Its computational description is challenging, especially if the temporal dynamics of the process are taken into account. 
Numerous methods to estimate saliency have been proposed in the last three decades. {They achieve almost perfect performance in estimating saliency at the pixel level, but the way they generate shifts in visual attention fully depends on winner-take-all (WTA) circuitry. WTA is implemented} by the biological hardware in order to select a location with maximum saliency, towards which to direct overt attention.
In this paper we propose  a gravitational model (GRAV) to describe the attentional shifts. Every single feature acts as an attractor and {the shifts are the result of the joint effects of the attractors}. 
In the current framework, the assumption of a single, centralized saliency map is no longer necessary, though still plausible. Quantitative results on two large image datasets show that this model predicts shifts more accurately than winner-take-all.
\end{abstract}
\begin{document}

\flushbottom
\maketitle
%
%
\thispagestyle{empty}

\section*{Introduction}

Despite the huge amount of data that reaches the human eye every second\cite{koch2006much},  neuronal hardware is insufficient to process it all at once. Indeed vision is. This crucial set of information is collected and forwarded to intermediate and higher levels of processing. The study of the visual attention mechanism has been in the spotlight for the past three decades~\cite{borji_2013}. It is at the crossroads of different disciplines such as psychology~\cite{smith2009integrated,hood1998adult}, cognitive neuroscience~\cite{duncan1998converging,martinez2013impact}, computer vision~\cite{itti1998,aim,Judd,zanca2017,sam,stateoftheart}. Although great advances have been produced, we are still far from defining a model that approximates human capabilities. Models of human visual attention are of great interest for the scientific community. They {help researchers to understand} the cognitive mechanisms of visual selection, which happens to be very intertwined with top-down processes~\cite{McMains2009,itti2001computational,connor2004visual,ZANCA2019}, but at the same time {they} provide a wide range of applications such as in marketing~\cite{hankinson2004brand,milosavljevic2012relative}, video compression~\cite{guo2009novel} or virtual reality~\cite{sitzmann2018saliency} {--} just to name a few. What makes modeling of human visual attention challenging is its inherently dynamic nature. Subsequent shifts in human attention are highly correlated with previous overt gaze shifts, as well as with the dynamics with which the scene itself changes~\cite{womelsdorf2006dynamic}, \zanca{the neural correlate of which, has been demonstrated to reside in a common population of neurons lying in frontal eye field (FEF)~\cite{corbetta1998common,nobre1997functional}.}

Current approaches in modeling human visual attention are based on the so-called saliency hypothesis~\cite{kochull}. It postulates the existence of a saliency map whose function is to guide attention and gaze towards the most conspicuous regions of the visual scene. This hypothesis has received numerous independent experimental confirmations~\cite{duan2015visual}. Following the approach traced by the seminal works of Itti and Koch~\cite{itti1998}, current approaches concentrate their efforts on the problem of learning saliency from human data. Attention models generally {yield} an output saliency map that indicates {the probability of fixating each location of the space}~\cite{stateoftheart,itti2005quantifying}. However, this does not model the temporal dynamics of the process in terms of the temporal order of fixations, i.e. scanpaths. For this reason, authors often assume that a circuitry~\cite{kochull,itti1998} of winner-take-all (WTA) is implemented by the human biological hardware to generate sequences of fixations, starting from saliency maps. This approach, however, still fails to provide a continuous dynamic of the process. Scanpath simulations are poor, not plausible and, consequently, not reliable for applications.

{In this paper,} we propose and validate a different model for generating attentional shifts over time. Our description is minimal both in the perceptual scheme and in the mathematical formulation. {In the proposed framework, the encoding of the oculomotor command needs a simple neuronal hardware, with the very mild assumption of units that perform sum operation, without the need of backward flows, leading to a plausible and straightforward} biological implementation.
The literature indicates that the V1 area is important for the conformation of the bottom-up saliency~\cite{zhang2012neural,olshausen1996emergence}, while other associative areas \zanca{such as V4, FEF and supplementary eye field (SEF)~\cite{westerberg2020priming}} receive signals simultaneously from both the V1 area and deeper layers~\cite{burkhalter1989organization}. Neurons in the V1 area can encode principal and independent components extracted directly from the visual input~\cite{olshausen1996emergence}{, and the} response magnitude of {such} neurons is greater when the stimulus is distinct from its surrounding~\cite{zhang2012neural,olshausen1996emergence}. {Moreover, some studies} confirm the effect of the V1 area on bottom-up visual attention in free-viewing scenes~\cite{duan2015visual}. {Following the outcome of scientific studies on the V1 area of the brain and its relationships in bottom-up visual attention,} we make the choice of minimal and naturalistic design, taking into account only basic features {to represent the input signal}, such as color, intensity and orientation gradients.

The main function of visual systems is to capture as much information as possible given limited resources. 
Its main sensory limitation is  spatial resolution which varies across the retina depending on the arrangement of photoreceptor mosaic and their receptive fields.
Under this considerations, we {provide} a mathematical formulation in which conspicuous features are modeled as masses that compete to attract visual attention. We derive laws that regulate attentional shifts through a gravitational model. The resulting shifts will be a consequence of {gravitational} attractions, together with a mechanism of inhibition of return (IOR), part of the visual foraging behavior,  that {allows the model to explore the whole scene}. While{, in principle,} such a description is {formulated independently from biological mechanisms, the} biological plausibility {of the proposed model} and a sketch of the neuronal hardware needed for realizing the underlying computation is given.

{The output of the model is a continuous function that describes the trajectory of the focus of attention.} It is worth {noting that the proposed framework makes it easy} to introduce additional visual features. External signals can be introduced to model the field of attention-grabbing masses to align it with specific tasks. Our approach relies on differential laws of motion, and it naturally provides the temporal dynamic of the exploration of the scene. 
Measures of scanpath similarity are adopted to measure plausibility of the model and closeness to real human data. In particular, three different metrics in literature have been shown to be robust: string-edit distance (SED)~\cite{Jurafsky,Brandt,Foulsham,zanca2019gravitational}, time-delay embeddings (TDE)~\cite{Wang} and scaled time delay-embeddings (STDE)~\cite{FixaTons,zanca2019gravitational}. 

\section*{Results}

\textbf{A gravitational model of visual attention.}
    A generic stream of visual input is defined on the domain $\mathbb{D}=\mathbb{R} \times \mathbb{T}\ ,$    where the subset $\mathbb{R} \subset \mathbb{R}^2$ represents the retina coordinates while $\mathbb{T} \subset  \mathbb{R}$ is the temporal domain.
    The visual attention scanpath is the trajectory $a(t):\ \mathbb{T} \rightarrow \mathbb{R}$, being $t \in \mathbb{T}$ the time index.
    \newrev{ Attention is assumed to be driven by the attraction triggered by a collection of $N$ {relevant visual features}. 
    Let $f_i:\mathbb{D} \to \mathbb{R}$
    be the function associated with the activation of a visual feature $i$ modeling the presence of a certain property in a pixel of the input stream,
    where $i \in \{1, ..., N \}$.}
    Larger values of $f_i(x,t)$ correspond with more evident presence of the visual feature in $(x,t) \in \mathbb{D}$, being $x$ the pixel coordinates.
    Let us assume to have the use of a number of $f_i$'s, each of them associated to different properties of the input stream.
    
    Inspired by the behaviour of gravitation fields, that naturally embed the idea of attraction, we model the visual attention scanpath as the motion of a unitary mass subject to the gravitational attraction of a distribution of masses $\mu$, associated to the visual features, $\mu: \mathbb{D} \to \mathbb{R} \ .$  In particular, we define $\mu(x,t) = \sum_{i}\mu_i(x,t)$, being $\mu_i$ the mass associated to feature $f_i$, that is
    $$\mu_i(x,t)= \alpha_i \lVert f_i(x,t) \rVert\ ,$$
    where the norm $\Vert \cdot \Vert$ measures the strength of the activation of $f_i$, and $\alpha_i > 0$ is a customizable scaling factor. 
    \newrev{ Notice that the $\alpha's$ values can properly be chosen to express the interest in a specific visual feature, thus providing task-driven trajectories.}
    We consider the gravitation field $E$, in which the attraction toward the distributional mass $\mu$ is inversely proportional to the squared distance from the focus of attention $a(t)$, given by
    \begin{equation}
    E(a(t),{t}) = - \frac{1}{2\pi} 
    \int_\mathbb{R} dx \frac{a(t)-x}{\lVert a(t)-x\lVert^2} \mu(x,t) := - (e * \mu)(a(t),t)  ,
    \label{overall_field}
    \end{equation} 
    where $*$ is the convolution operator and $e(z) = (2\pi)^{-1}(z) \Vert z \Vert ^{-2}$.
    Once we are given the gravitational field, we can compute the {Newtonian differential equation}, that are
    \begin{align}
    \ddot a(t) + \lambda \dot a(t) + (e \ast \mu)(a(t),{t}) = 0,
    \label{Trajectory}
    \end{align}
    where dumping term $\lambda \dot a(t)$, with $\lambda > 0$, prevents from oscillations typical of gravitational systems and it helps to produce precise ballistic movements toward the salient target. 
    Integrating Eq.~\ref{Trajectory} allows us to compute the visual attention trajectory at each time instant.\footnote{We converted the equation to a first-order system of differential equations, as commonly done, introducing auxiliary variables. Then we used the \texttt{odeint} function of the Python SciPy library, in the setting in which it automatically determines where the problem is stiff and it chooses the appropriate integration method.}

    The choice of the visual features that induce the corresponding masses is determinant in modeling the behaviour of the attention system. A key property of the proposed model is that there are no restrictions on the categories of features one could {exploit}. While basic low-level features are considered in this work, other features associated to semantic categories (faces, objects, actions, etc.)  could be introduced that might be relevant in specific visual exploration tasks.
    {In particular, the} features we consider in this paper are described as follows.
    \begin{itemize}
    	\item Let $i:\mathbb{D} \to \mathbb{R}$ be 
    	  the intensity of the frame, that yields the feature associated to \textit{spatial gradient of the brightness},
    	$\mela{f_1} = \nabla_x i $. 
    	This features carries information about edges and, generally speaking, 
    	it reveals the presence of details in the input data.
    	
    	\item Let $c_j:\mathbb{D} \to \mathbb{R}$ be 
    	  the color channels of the frame, with $j \in \{1, 2, 3\}$ that yields the feature associated to \textit{spatial gradient of the color},
    	$f_{1+j} = \nabla_x c_j $. 
    	This features carries information about edges {on the color channels and, similarly to the case of $f_1$,}
    	it reveals the presence of details in the input data.
    	\item  Let $o_k: \mathbb{D} \to \mathbb{R}$ be the \textit{orientations}, that reveal the presence of edges oriented at $0$, $45$, $90$ and $135$ degrees, with $k \in \{1, 2, 3, 4\}$ . The feature 
    	${f_{4+k}} = \nabla_x o_k $ characterizes  areas oriented in a certain direction.
    \end{itemize}
    
    Please notice that here we use a natural assumption of describing the input stream with contrastive features given by the gradient function. It aims at reproducing the activity of photoreceptive cells working in color-opponent. In humans,  after a reflexive shift of attention towards the source of stimulation, there is an inhibition to remain in the same location. This mechanism is called Inhibition Of Return (IOR). \zanca{Originally discovered in human studies of attention, inhibition of return is a tendency for the organism to orient away from a previously attended location and biologically depends on neural structures that participate in oculomotor control~\cite{westerberg2020priming}, parietal and frontal cortex~\cite{bichot2002priming}. This phenomenon was first described by Posner and Cohen~\cite{posner1985} who showed that reaction times to detect objects appearing in previously cued locations were longer than to uncued locations. The phenomenon has been demonstrated in a number of different paradigms~\cite{gibson1994inhibition,pratt1999inhibition} as part of the visual foraging behavior and may reflect a \textit{novelty bias}~\cite{milliken1998attention} making visual exploration to proceed efficiently. Other authors have suggested alternative explanations of IOR depending upon task specificity such as the abservation of an inhibitory effects for non-spatial attributes of irrelevant pre-cues~\cite{mondor1998inhibitory,law1995color}, or computational models of negative priming~\cite{houghton1984model,milliken2000attending} that can also generate IOR from irrelevant cues. While other implementations are equivalent and compatible with the present proposal, here we have made the choice to design the IOR in its original description, adapting it to the gravitational framework.} We define a similar mechanism in our model, to prevent the trajectory to get trapped into regions of equilibrium and favour complete exploration of the scene. The dynamic of a function of inhibition $I(x,t)$ can be modeled as  
    \begin{equation}
    \frac{\partial I(x,t)}{\partial t} + \beta I(x,t) = \beta g(x-a(t)),
    \label{IoR-Eq}
    \end{equation}
    where $ g(u) =  e^{- \frac{u^{2}}{2 \sigma^{2}}}$ and $0 < \beta <1$. This is directly applied to the feature masses, in order to decrease the gravitational contribution from already-visited spatial locations. As a results, the distribution of masses $\mu$ becomes
    \begin{align}
    \mu(x,t)=\sum_i \left( \mu_{i}(x,t) - \mu_{i}(x,t) I(x,t)) \right) .
    \label{OverallMass}
    \end{align}
    
	\textbf{Scanpath prediction.} 
        Shifts on visual attention allow mammals to relocate the fovea to the next location of interest. A sequence of shifts determines a visual scanpath. 
        \zanca{The gravitational model described above provides a computational method to produce sequences of fixations and saccades, given a visual stimulus. Equation~\ref{Trajectory} provides a differential law describing attentional shifts. It can be numerically integrated to produce simulations that can be compared with data from human subjects collected by means of eye-trackers. It is worth mentioning that our model generates a continuous scanpath. The same fixation detection algorithms that are used on human recordings have been applied here to extract fixations from the output of the gravitational model. Instead, the WTA is an algorithm that provides a discrete output. }
        
        \zanca{The comparison between the two models is based on similarity metrics or distances between trajectories. These metrics quantify how well the simulated sequence fit the locations visited by the human subject, taking into account also the order in which these locations are visited. In particular, results are given in terms of the following metrics:
        \begin{itemize}
            \item \textit{String-edit distance (SED)~\cite{Foulsham,zanca2019gravitational}}. The
input stimulus is divided into
$m\times m$ regions, labeled with characters. Scanpaths are
turned into strings by associating each fixation with the corresponding region. Finally, the string-edit algorithm~\cite{Jurafsky} is used to provide a measure of the distance between the two generated strings.
            \item \textit{Time-delay embeddings (TDE)~\cite{Wang}.} This measure is commonly used in order to
quantitatively compare stochastic and dynamic scanpaths of varied lengths. It is defined as the average of the minimum Euclidean distances of each sub-sequence of length $m$ from the original trajectory with all the possible subsequences of length $m$ from the generated trajectory.
            \item \textit{Scaled time-delay embeddings (STDE)~\cite{FixaTons,zanca2019gravitational}.} This scaled version of the previous metric is obtained by normalizing coordinates between 0 and 1, according to the size in pixels of each of the presented stimuli. 
        \end{itemize}}
        {The results in Table \ref{tab:datasets1} and Table \ref{tab:datasets2} summarized} the scores {(SED, TDE and STDE metrics)} calculated with respect to the human scanpaths. 
        \zanca{Both models under examination, i.e. GRAV and WTA, assume that pre-attentive spatial maps are given in the system. In the comparison we include two possibilities to ensure a fair comparison. On the one hand, we follow the original implementation of the WTA and assume that a saliency map is pre-calculated and fed as input to the systems. We use the original implementation described in Itti~\cite{itti1998}. In the second case, we use a set of more basic features, corresponding to intensity, color and orientation.}
        
        GRAV model outclasses the WTA in all cases. The results   show that the introduction of the saliency map does not bring significant benefits to either model. We hypothesize that the advantage of GRAV over WTA depends on the fact that the gravitational approach allows to generate more naturalistic fixations, more centered on the center of mass of salient objects rather than drastically on the edges. This is also shown qualitatively in the figure~\ref{fig:scanpaths}.

        \begin{table}[h]
        	\begin{center}
        		\begin{tabular}{|p{.1\linewidth}|p{.18\linewidth}|p{.13\linewidth}|p{.13\textwidth}|p{.13\linewidth}|}
        			\hline
            			\textbf{Model} & \textbf{Pre-attentive maps} & \textbf{SED} & \textbf{TDE} & \textbf{STDE} \\
        			\hline
            			{GRAV} & {Basic} & {7.68 (0.65)} & \textbf{226.70 (76.96)} & \textbf{0.80 (0.06)} \\
        			\hline
            			{GRAV} & {Itti} & \textbf{7.67 (0.63)} & {228.08 (76.97)} & {0.79 (0.06)} \\
        			\hline
            			{WTA} & {Basic} & {8.41 (0.50)} & {425.27 (66.87)} & {0.65 (0.04)} \\
        			\hline
            			{WTA} & {Itti} & {8.41 (0.49)} & {417.12 (65.99)} & {0.66 (0.04)} \\
        			\hline
        		\end{tabular}
        		\caption{\textbf{Results on MIT1003.} In bold, the best results on average. The standard deviation values are given in round brackets. Note that the two models have performance equivalent to varying basic features. The  gravitational model performs better on every metric, compared to the winnner-take-all model. }
        		 \label{tab:datasets1}
        	\end{center}
        \end{table}

        \begin{table}[h]
        	\begin{center}
        		\begin{tabular}{|p{.1\linewidth}|p{.18\linewidth}|p{.13\linewidth}|p{.13\textwidth}|p{.13\linewidth}|}
        			\hline
            			\textbf{Model} & \textbf{Pre-attentive maps} & \textbf{SED} & \textbf{TDE} & \textbf{STDE} \\
        			\hline
            			{GRAV} & {Basic} & {13.81 (2.01)} & \textbf{454.52 (111.17)} & \textbf{0.78 (0.04)} \\
        			\hline
            			{GRAV} & {Itti} & \textbf{13.77 (2.01)} & {458.76 (110.79)} & \textbf{0.78 (0.04)} \\
        			\hline
            			{WTA} & {Basic} & {14.48 (2.07)} & {762.99 (100.94)} & {0.66 (0.03)} \\
        			\hline
            			{WTA} & {Itti} & {14.48 (2.07)} & {766.06) (101.92)} & {0.66 (0.03)} \\
        			\hline
        		\end{tabular}
        		\caption{\textbf{Results on CAT2000.} In bold, the best results on average. The standard deviation values are given in round brackets. }
        		 \label{tab:datasets2}
        	\end{center}
        \end{table}

\section*{Discussion}

    In the literature, with the influence of Treisman and Gelade's feature integration theory\cite{treisman_gelade} and after the seminal work by Koch and Ullman\cite{kochull} and Itti et al.\cite{itti1998}, attention models are often associated with the estimation of saliency maps. {Benchmarks} of  saliency prediction are well established\cite{mit-saliency-benchmark}. Less studied is the problem of generating fixation sequences, along with the problem of explaining whether these fixations actually depend on a previous computation to combine basic features into a saliency map. In fact, such a computation seems non-trivial, albeit plausible biologically\cite{kochull}. 
    Computational models of saliency often tacitly assume that fixations can be generated with the winner-take-all mechanism\cite{kochull} along with some unspecified rules of preference. Other methods for the prediction of visual attention shift have been proposed in the literature. They all assume that shift of visual attention are based on features extracted in a pre-attentive phase. In~\cite{le2016introducing} the authors incorporate in the model a series of biological biases that allow for more plausible saccades, on the top of a  saliency map. Even though this method delivers more precise saliency estimate, it fails to provide an explanation of the phenomenon, i.e. how these biological biases actually emerge. Other authors have developed different theories, independent of saliency (but still assuming the existence of spatial maps of features extracted in parallel in a pre-attentive phase) but their descriptions are only qualitative. This often does not allow a description of the computation underlying the visual system and, consequently, prevents a quantitative comparison. For example, Renniger et al.~\cite{renninger2005information} provides an explanation of how humans explore specific artificial shapes, but it is not clear how this can be extended to a general theory of attention. Recently, data driven machine learning approaches tried to predict sequences of human fixations directly from data~\cite{jiang2016learning, kummerer2017deepgaze}. These approaches have two main flaws. The first is that their performance depends heavily on the data chosen. The second is that they fail to give a computational description of biology.  While from an application point of view they can be very useful, they say little about how human perception works. It is also unclear how these models can extend to the natural case of dynamic scenes.
    
\begin{figure}
        \begin{minipage}{.4\textwidth}
          \centering
          \includegraphics[width=.4\linewidth]{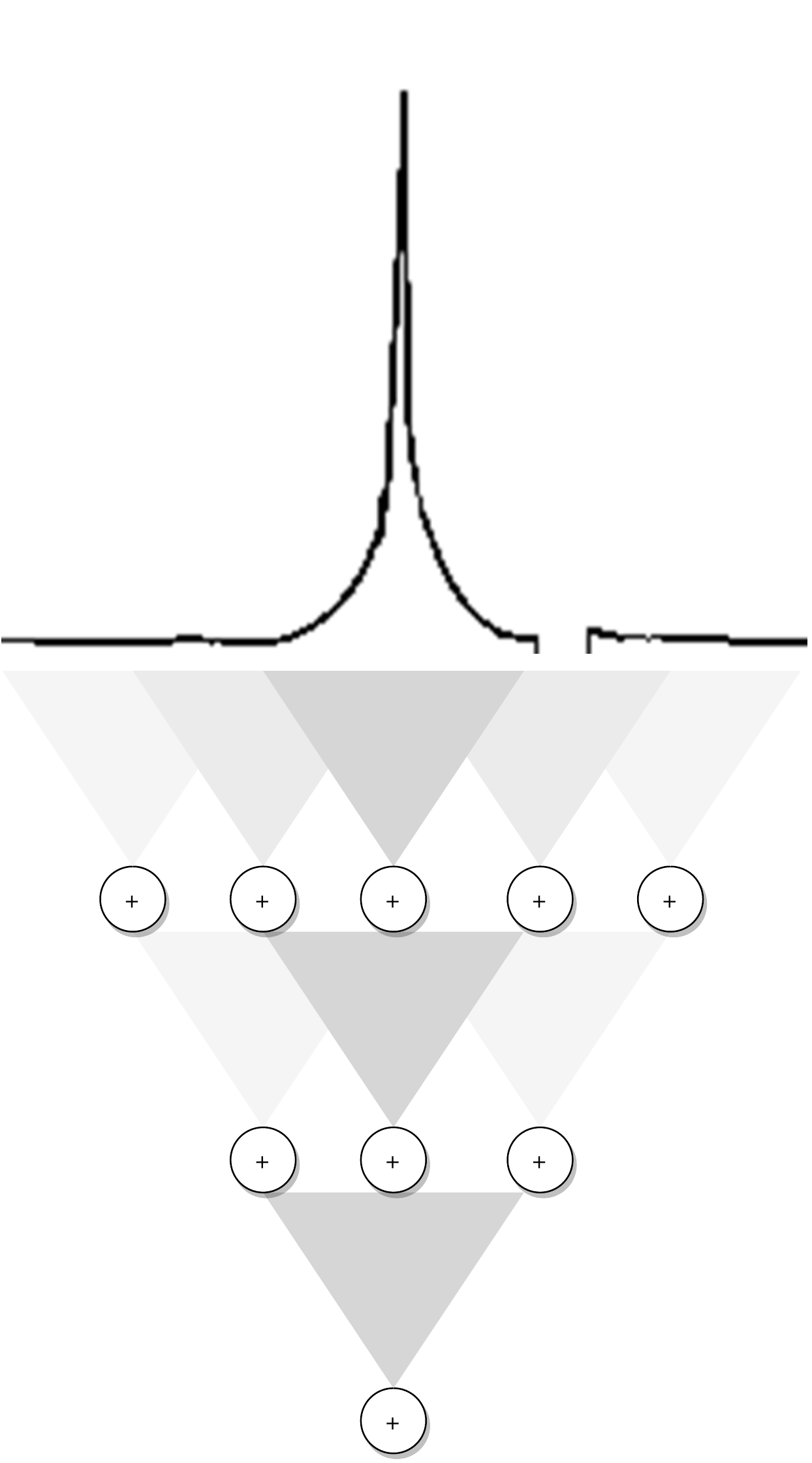}
                \caption{Field network. This network realizes the computation of a quantity proportional to the functional associate with the gravitational field. The black line on the top is illustrative. It shows a qualitative example of the distribution of cones in the retina. The maximum point correspond to the center of the fovea. A characteristic blind spot is also illustrated.}
                \label{fig:fieldnetwork}
        \end{minipage}
        \begin{minipage}{.6\textwidth}
          \centering
          \includegraphics[width=\linewidth]{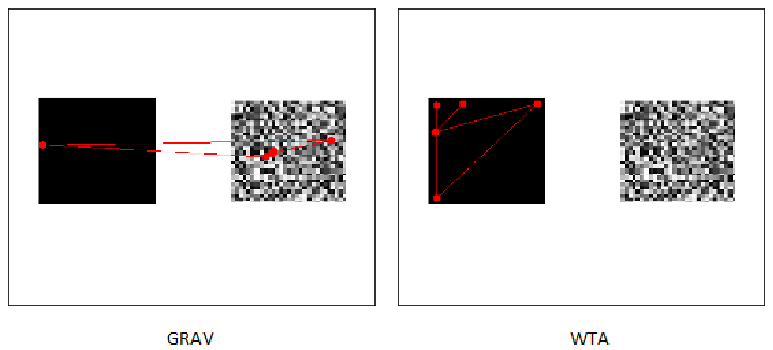}
                \caption{Example of simulated scanpath. This example shows a borderline case where the scanpath generated with WTA is unnatural because it focuses exclusively on borders with high center-surround differences. The GRAV approach, in contrast, allows to generate fixations on center of mass. Consequently, the large amount of variation in random noise on the right makes it more interesting than the square on the left.}
                \label{fig:scanpaths}
        \end{minipage}%
    \end{figure}

    The proposed gravitational approach GRAV allows to explain attentional shifts better than the current reference model, i.e. WTA. 
    With the same features, the proposed model outperforms the winner-take-all algorithm in the task of scanpath prediction. The result is independent of the choice of the starting features, which we assume to be calculated in parallel in a pre-attentive phase. The results are more evident in the case of TDE and STDE measures, which are based on Euclidean metrics. These are, in fact, more spatially sensitive and could capture any recurring dynamics~\cite{abarbanel1994predicting}, such as the preference for shorter saccades. 
    As the figure~\ref{fig:scanpaths} shows qualitatively, both the distribution of the amplitudes of the saccades but also the presence of fixations in the center of objects. This claim has been quantitatively demonstrated with metrics to measure the proximity of scanpath measured with human ones. 
    \zanca{Unlike WTA, GRAV models do not strictly relay on a pre-calculated saliency map. It acts directly on feature maps that are treated as mass distributions. This ensures the versatility of the model and could directly explain how other priors, i.e. top-down priors, can intertwine in the attention mechanism, as long as they preserve the spatial conformation. 
    \newrev{Clearly, it is not the case that the eyes are purely driven by low level feature changes.  
    Even in the absence of an explicit task, other factors such as \textit{meaningfulness} of a location in a scene~\cite{jhend} can  predict  fixations. 
    Similarly, it has been shown that people exhibit  an  understanding of \textit{scene grammar} and move their eyes correspondingly~\cite{vo2019reading}, 
    thanks to the interpretation of the scene.  
    A preliminary attempt to interpret these high-level visual skills has been already reported in~\cite{ZANCA2019}, where it is shown 
    how the hidden neurons of a deep neural network trained for object classification can be integrated with a variational model, 
    provided that  the spatial map distribution is maintained .} 
    Furthermore, the GRAV model describes a continuous dynamics of the process. The output of the model presents the same step behavior of saccades that is determined by the joint contribution of the inhibition mechanism. This partially explains the advantage of the GRAV model over the WTA in terms of performance in the scanpath prediction task: when performing a saccade, in fact, the gravitational contribution of the peripheral vision continuously  influences the relative positioning with respect to the final target~\cite{veneri2011spike}.}
    Describing a computational model of visual attention while taking into account how this process could be implemented by a biological hardware requires considering that visual attention solves the sensory-functional trade-off  between minimizing resources and organizing them efficiently to collect information, in a hierarchical structure.
    From this point of view, the proposed differential model has a natural interpretation in terms of local computation made by a hierarchical layers structure  in which each unit is identified by a layer index $l \in L$ and a positional index $i \in \{1, ..., n_l\}$, where $n_l$ is the number of units belonging to the $l$\textit{-th} layer. The first layer, $l_0=0$, is the system input and its units can be identified with the photoreceptors distributed on the retina. Then, $\forall l \in L - \{l_0\}$, units are defined by $u_i^l = \sum_{j \in N_i^l} \sigma(u_j^{l-1}),$ where $N_i^l$ represents the receptive field of the unit $u_i^l$ and $\sigma$ is an activation function which eventually introduce non-linearity in the computational graph. We identify two feed-forward steps for the calculation of the feedback signal encoding the eye motion command (which, in our description, is a continuous signal). This steps are, respectively, the calculation of the quantity in equation~\ref{overall_field} (associated with the gravitational field) and the updating rule of the variable $\ddot a(t)$ (encoding the eye shifts) which derives from the differential equation~\ref{Trajectory}. The first step is realized by a hierarchical structure (see figure~\ref{fig:fieldnetwork}) in which a layer of computational units perform a linear summation with equal weights to achieve an isotropic response. In other words, the activation function $\sigma$ is linear: units at a layer $l$ receive the activations in a neighborhood  in the $l-1$ layer and propagate to the $l+1$ layer an amount of activation which is proportional to the linear summation. We will call it \textit{field network}. Note that calculating a linear summation response with saturating non-linearities on the inputs requires \textit{ad hoc} adjustment of the connection weights, depending on the activity intensity and number afferent cells~\cite{hmax}.
    It is well known that the visual system is organized retinotopically and hierarchically from the retina to the visual cortex V1. The central part of the retina, the fovea, is an area of $2.5$ degrees with the best visual resolution (capacity to recognize fine details); outside this area, spatial resolution decreases sharply.  In fact, while receptive field (RF) of neurons corresponding to the fovea are small, their density is high and they are overrepresented in V1, the size of RF increase and their number decrease with eccentricity. Furthermore, units of the periphery project diffusely to many central neurons which receive information from wide receptive fields~\cite{carpenter}. The effectiveness of this neuronal organization of resources furthermore results on an enhanced visual system's effective spatial resolution~\cite{carrasco}. The representation of stimuli with a peak function naturally implements the weighting term $\frac{a(t)-x}{\lVert a(t)-x\lVert^2} \propto \frac{1}{\lVert a(t) - x \lVert}$ within the functional action, which in that equation had a gravitational interpretation.  In fact, the units connected to the receptors closer to the central retina will receive with a probability related to their distribution a greater amount of activation and, consequently, will propagate a stronger signal. 
    
    \zanca{We have discussed the gravitational computations at the levels of description at which the nature of the computation itself is expressed (i.e., mathematical analysis) and at which the algorithms that implement the hierarchical computation are characterized. Now we provide a sketch of  the neuronal hardware~\cite{marr1976understanding} that may be implementing this scheme for visual attention shifts based on attractor's laws. It is well known that simple properties are extracted from the retina to the early representation, corresponding to basic features. This representation are most likely localized within and beyond striate cortex, like the V4 for color and geometric shapes, or middle temporal and middle superior temporal areas for motion. This spatial maps are fed into the hierarchical structures and provide the input for the GRAV network. 
    As it was proposed in the original paper of the WTA~\cite{koch1987shifts}, a supported hypothesis is that this computations may take place in the early visual system, for example in the lateral geniculate nucleus LGN. There is in fact evidence that visual signals can  travel from the periphery to the cortex and back to the LGN~\cite{briggs2007fast}. This hypothesys would be more in line with recent findings confirming that attention modulates visual signals before they even reach cortex by increasing responses of both magnocellular and parvocellular neurons in the LGN~\cite{mcalonan2008guarding}.}
    
    It is worth underlying at this step that our algorithmic formulation offers a simplified explanation of the same phenomenon, compared to the WTA circuitry described by Koch and Ullman and their proposal for a biological implementation~\cite{kochull}. GRAV requires elementary units to perform (weighted) linear summation, while Koch and Ullman assume (i) units to perform a max operation and (ii) hypothesize the presence of a parallel network, identical in structure but performing backward calculations. The latter is introduced into their framework as a trick to retrieve the spatial location of the maximum. It is not specified, then, how this is encoded and transmitted to subsequent layers for further calculations and to generate a command signal for relocation of the fovea. In our gravitational framework, no backward computation are required. In fact, the quantity calculated by the field network will be used \textit{as it is} to update a feedback signal that codify the fovea shift. The whole stack of computation is feed-forward and include only local computation, which makes the overall process more efficient. The neural implementation of the second step of computation in equation~\ref{Trajectory} requires the definition of a neuronal graph implementing an integration of the given differential equations~\ref{Trajectory}. It is well known that a neural network implementation of differential equation is possible~\cite{lee1990neural,lagaris1998artificial,tsoulos2009solving} and efficient in terms of execution time, compared with classical method that do not exploit parallel computing~\cite{Yadav2015}. In particular, a simple implementation of finite differences methods is presented by Lee and Kung~\cite{lee1990neural} together with an explicit method for calculating a general continuous and discrete neural algorithms for solving a wide range of complex partial differential equations. This scheme {assumes} basic functional operation that are plausibly implementable by a biological neural circuit. However, the implemetation of this second step become straightforward assuming that the dissipation term introduced in the theoretical model could be solely reduced to phenomena of friction. This derives from the purely mechanical fact of the eye residing in a plant of muscles that keep it in its natural position of \textit{looking straight}. The  dissipation term is fundamental in the proposed  model to ensure precise moviments toward a target and finds its biological counterpart in the   resistance  to movement that derives from the plant in which the eye is placed. In this case, the updating equation would be dramatically simplified to  $ \ddot a(t) \propto - (e \ast \mu)(a(t),{t}),$   obtaining that the output of the field network on the first step directly codify a speed command to be sent to the eye muscles. Also in this case, and differently by the Koch and Ullman's proposal~\cite{kochull}, no backward signal is necessary which gives rise a more natural implementation and efficient computing.
    
    \newrev{Finally, we notice that the dynamic nature of the model is particularly suitable for virtual reality applications.
    Without any modification, the proposed model can be used to navigate 360-degree environments. 
    This could open the doors to a new research direction, where we emphasise the reproduction of conditions that are increasingly 
    similar to human vision.}

\section*{Methods}

    \textbf{Datasets.} Collecting eye-tracking data is a time-consuming process. Selected subjects must be invited to participate in an experiment that normally takes place at the same room, with controlled light conditions to limit the variability of the experiment, {and with the need of calibrating} the eye-tracking tools. In recent years, large collections of data have been made publicly available. Due to the inherently complex nature of both the stimuli and the human cognitive process, bigger eye-tracking data are necessary for a {meaningful} evaluation. For this reason, through all experimental evaluations {of this paper}, we use 2 different publicly available eye-tracking datasets. The exposure time of subjects to visual input ranges from 3 to 5 seconds. The number of subjects per stimulus varies from 15 to 24. Image resolution varies widely within the dataset. The details for each of the two datasets used are specified in Table~\ref{tab:datasets}.
    All images and video frames are resized to a resolution of 224x224. {This makes the experiments more easily manageable, reducing the computational time}. We noticed that higher resolutions did not significantly improve the performance of any of the models.
        \begin{table}[h]
        	\begin{center}
        		\begin{tabular}{|p{.15\linewidth}|p{.8\textwidth}|}
        			\hline
            			\textbf{Dataset name} & \textbf{Details} \\
        			\hline
            			{MIT1003}~\cite{Judd} & This dataset contains 1003 natural indoor and outdoor scenes. They are sampled  with variable sizes, where each dimension is in ${[405,1024]}$px. The database contains 779 landscape images and 228 portrait images. Fixations of 15 human subjects are provided for 3 seconds of free-viewing observation.	 \\
        			\hline
            			{CAT2000}~\cite{FixaTons} & A collection of 2000 images is provided {as} the training portion of this dataset. Semantic content {largely} vary among twenty different categories. The resolution of the images is $1920\times 1080$px. Fixations of 24 human subjects are provided for 5 seconds of free-viewing observation.\\
        			\hline
        		\end{tabular}
        		\caption{\textbf{Collection of datasets.} To ensure a proper evaluation of the proposed model, a large collection of static images from two different datasets have been used. Eye-tracking data is collected in a free-viewing setup. Details are described in the right column for each of the datasets.}
        		 \label{tab:datasets}
        	\end{center}
        \end{table}

	\textbf{Feature map extraction and software implementation.} The proposed gravitational model is compared with the WTA algorithm in the task of generating fixation sequences (equiv. scanpath) on a large collection of static images. Although the gravitational model has the advantage of being naturally extended to dynamic scenes (i.e., video{s}), we restrict ourselves to static images {to make the comparison easier to evaluate}. The effectiveness {the gravitational model, named GRAV,} in predicting saliency and scanpath on dynamic scenes has been demonstrated in a previous work~\cite{zanca2019gravitational}. 
	We use the same input features proposed in the computational implementation of the WTA model realized by Itti~\cite{itti1998}. Such features include an intensity channel, three color channels and four orientation channels. Notice that all the {feature} maps are equally weighted and no {special tuning} is applied, in any case, that could have improved the performance of the gravitational model. Since in its original version~\cite{kochull} WTA does not {directly} work on {such} basic features but on a saliency map obtained with subsequent calculation steps, both models are also evaluated while operating on the saliency map, instead of the basic features. The saliency map is generated using the code in {the} original implementation provided by the authors{, and the} implementation of the WTA model follows the description in the original paper~\cite{kochull}. The first fixation is chosen as the location with maximum saliency value. In the case of basic feature maps, they are first combined linearly with equal weight, then the location with maximum value on the resulting map is selected. {Moreover}, the selected location is inhibited within 2 degrees of visual angle to switch attention to a subsequent location. In the original paper~\cite{kochull}, the authors propose two additional rules for selecting subsequent locations based on proximity and similarity preference. However, they do not provide quantitative descriptions of how this should be implemented either mathematically or by biological hardware. For this reason, these rules have not been implemented in our software. It is worth pointing out that the concept of proximity preference is, instead, automatically encoded in the GRAV model that we propose in this paper. It derives from its gravitational description, where attraction is inversely proportional to the distance. More details about biological reason of such a resulting behaviour will be given in the following paragraphs.

    \textbf{Tuning of the parameters.}  The behaviour of the proposed differential model {GRAV} depends on a small set of parameters $\{ \beta,\lambda \}$ that must be carefully selected. 
    The parameter $\beta$ was set to $0.1$. We found that the choice of different values for $\beta$ did not produce significantly different results. 
 The parameter $\lambda > 0$ prevents oscillations (or orbits) because it introduces a dumping term (see, for example, the classic equation for the damped harmonic oscillator).
    For all the experiments {we used the set of parameters that maximized the performance of GRAV in 20 images that were kept out from the described datasets for validation purposes}. In particular, we {performed a grid search procedure, selecting the parameters that maximized} the NSS saliency score.

\bibliography{references}

\section*{Acknowledgements}
We thank Fr\'ed\'eric Precioso and Lucile Sassatelli for fruitful discussions on the model which stimulated a wider view on the topic as well as interesting application perspectives in computer vision.

\section*{Author contributions statement}
D. Z., M. G. and S. M. contributed to the mathematical formulation of the model. D. Z. and A. R. contributed to the analysis of the biological plausibility. D. Z. conceived and conducted all the experiments. All authors analysed the results and reviewed the manuscript.

\section*{Additional information}

Accession codes: All codes used for the experiments reported in this manuscript will be made available in a public repository.
Competing Interests: The authors declare no competing interests. The work was partially supported by RoNeuro and Liquidweb srl who participate in the Neurosense Joint Lab of the Dept. Medicine Surgery and Neuroscience, University of Siena.

\end{document}